\documentclass[sigconf,nonacm]{acmart}
\makeatletter
\@ACM@balancefalse
\makeatother
\usepackage{amsmath,booktabs,graphicx,microtype}

\usepackage{algorithm}
\usepackage{algpseudocode}

\usepackage{tikz}
\usetikzlibrary{arrows.meta,positioning,fit,backgrounds,shapes.geometric}
\usepackage{cleveref}
\usepackage{xurl}
\usepackage{listings}
\usepackage{xspace}
\usepackage{enumitem}
\usepackage[most]{tcolorbox}
\newtcolorbox{summarybox}{
  colback=black!5, colframe=black!45, boxrule=0.4pt, arc=2pt,
  left=5pt, right=5pt, top=4pt, bottom=4pt, breakable,
}

\usepackage{tabularx} 

\usepackage{multirow} 

\newcommand{\ourmethod}{\textsc{{GAD}}\xspace}
\newcommand{\affMSU}{%
  \affiliation{%
    \institution{Michigan State University}
    \city{East Lansing}
    \country{USA}}}
\newcommand{\affBaidu}{%
  \affiliation{%
    \institution{Baidu, Inc.}
    \city{Beijing}
    \country{China}}}
\newcommand{\affMIT}{%
  \affiliation{%
    \institution{Massachusetts Institute of Technology}
    \city{Cambridge}
    \country{USA}}} 
\begin{document}
\title{Goal-Aware Dynamic Multi-Agent Collaboration for Long-Horizon Tasks}
\author{Xu Shen}
\affMSU
\email{shenxu@msu.edu}

\author{Lingyong Yan}
\affBaidu

\author{Shuaiqiang Wang}
\affBaidu

\author{Jianmin Wu}
\affBaidu

\author{Zhikai Chen}
\affMSU
\author{Jiliang Tang}
\affMSU
\author{Jie Ren}
\authornote{Corresponding author.}
\affMIT
\email{rjie@mit.edu}
\begin{abstract}
LLM-based agents are increasingly used to complete complex, long-horizon tasks without human intervention. To make progress toward the task goal, agents need to perform different duties, such as planning and verification, when needed. To understand how single agents manage these duties during long-horizon execution, we conduct a preliminary study. Our analysis shows that verification has the strongest association with task success, but often occurs late, leaving limited opportunities for correction. We also find that the proportion of duty-relevant information decreases as records from different duties accumulate in a shared context. These findings lead us to examine whether multi-agent systems (MAS) can address these problems by assigning duties to agents with separate contexts. Our analysis shows that this separation alone is insufficient. Agents can lose track of the overall goal and repeat prior investigations. They may also lack necessary background while receiving irrelevant information from others. To maintain goal awareness and coordinate collaboration as task needs change, we propose \textbf{G}oal-\textbf{A}ware \textbf{D}ynamic Multi-Agent Collaboration (\ourmethod{}). It uses a growing execution graph to organize subtasks and their dependencies. To keep agents focused on their goals, each context retains the original instruction and the assigned task's goal and acceptance conditions. To adapt collaboration and information sharing during execution, agents can divide tasks into subtasks and specify dependencies based on execution evidence. These dependencies guide when agents work and which results they receive. Verification and recovery check subtask results and full task completion against their goals, providing feedback for repair when checks fail. Experiments with Claude Sonnet~4.6 and Qwen3.7-Plus on Terminal-Bench~2.1 and SWE-bench Pro Hard-51 show performance improvements over the evaluated baselines and lower token costs than multi-agent baselines on jointly solved tasks.
\end{abstract}
\maketitle
\raggedbottom

\section{Introduction}






LLM-based agents have shown promising capabilities in completing complex tasks without human intervention, particularly on long-horizon tasks~\cite{li2026loopsbench,wang2026looparena}. These tasks can span hundreds of interaction steps, during which agents repeatedly interact with the environment and adapt their actions based on feedback~\cite{li2026longhorizonterminalbenchtestinglimitsagents,ma2026longhorizon}. To guide this process toward successful completion, agents often need to perform various \emph{duties}, such as planning, reasoning, verification, and functional responsibilities~\cite{erdogan2025planandact,jin2025searchr1,asai2024self}, as needed throughout execution. For example, a web research agent first plans what information to search for~\cite{li2026webthinker}, then reasons over the retrieved evidence~\cite{jin2025searchr1}, before verifying whether the evidence supports its answer~\cite{asai2024self}. Analysing these duties is crucial to understand agent trajectories. If an agent cannot allocate and switch duties when needed, it may continue executing outdated plans or leave errors unchecked~\cite{ma2026longhorizon}. Therefore, it is important to understand how agents manage these duties during long-horizon execution.

However, allocating duties effectively by a single agent can be challenging. In Section~\ref{sec:analysis-single}, we perform a preliminary study to examine how agents manage these duties. Our analysis reveals two key limitations: \textbf{(1)} Among all duties, verification exhibits the strongest association with task success, yet often occurs late in the task process. Intuitively, early verification of intermediate results can help agents detect and correct errors before they propagate into subsequent actions. However, agents may continue working on the task without performing the duty of verification, leaving errors undetected and limiting opportunities for correction. \textbf{(2)} Agents may struggle to retrieve relevant evidence and prior decisions for the current duty from the extensive accumulated context. Our context analysis shows that as records from different duties accumulate in a shared context, information relevant to the current duty accounts for a decreasing proportion of the interaction history. These two findings motivate us to investigate whether explicitly assigning distinct duties to separate agents using multi-agent systems (MAS) can address these limitations by maintaining separate contexts and enabling verification when needed.

However, the naive adoption of MAS does not resolve the limitations of single-agent execution. In Section~\ref{sec:analysis-mas}, we evaluate six representative multi-agent configurations on duty performance using the same trajectory analysis approach. Our analysis identifies two limitations. \textbf{(1)} Agents can lose track of the overall goal. They might repeat investigations that others have already attempted, spending additional steps without advancing the task. \textbf{(2)} Separate contexts do not automatically improve information management. Agents may lack the background or conditions needed for their duties, while messages from other agents may add irrelevant information to their contexts. These findings highlight that throughout long-horizon execution, maintaining goal awareness and dynamically coordinating duties and information exchange are important.

Motivated by these findings, we propose \textbf{G}oal-\textbf{A}ware \textbf{D}ynamic Multi-Agent Collaboration (\ourmethod{}) for long-horizon tasks. GAD uses a growing execution graph to organize subtasks and guide information sharing during execution. Each node represents an assigned task, while edges record task decomposition and dependencies. To prevent goal drifting, we retain the original instruction and the assigned task’s goal and acceptance conditions in each agent’s context. Each task also retains its goal and the duty to satisfy it as the graph expands. To support dynamic collaboration, agents can divide their assigned tasks into subtasks and specify dependencies based on execution evidence. These dependencies guide agent scheduling and dependency-aware context construction to supply relevant results. Verification and recovery check both subtask results and full task completion against their goals and acceptance conditions, providing feedback for repair when checks fail. We conduct experiments on Terminal-Bench~2.1 and SWE-bench Pro Hard-51 with Claude Sonnet~4.6 and Qwen3.7-Plus, demonstrating performance improvements over the evaluated baselines. Further analyses support component effectiveness and show lower token costs than multi-agent baselines on jointly solved tasks.




\section{Related Work}

\subsection{Long-Horizon Agents}
Recent research improves long-horizon agents through supervised fine-tuning and multi-turn reinforcement learning. The former trains agents on synthetic interaction trajectories ~\cite{xu2025agenttrek,yang2026swe,zhu2026termigen},
while the latter develops decision-making policies through feedback from real or synthesised environments
~\cite{wang2025ragen,xi2025agentgym,chen2026scaling,wang2026agent}. These approaches improve agent behaviour through training, but extended execution also requires managing accumulated information and tracking task progress. MEM1 jointly learns reasoning and compact memory updates ~\cite{zhou2026mem1}.
ReSum summarises interaction histories to support continued search~\cite{wu2025resum}, while Context-Folding trains agents to handle subtasks in separate trajectories and retain their summaries~\cite{sun2025scaling}.
LongHorizon-Harness further separates task management,
execution, and auditing to maintain independently checked progress~\cite{ma2026longhorizon}.
Our method complements policy training by organising collaboration during execution. It connects agent-proposed subtasks, dependency-guided information sharing, and verification through a growing execution graph, while retaining the task goal.
\subsection{Multi-Agent Systems}
Multi-agent collaboration has developed from predefined role procedures towards automatically constructed collaboration structures~\cite{hong2024metagpt}.
GPTSwarm optimises agent prompts and graph connections
~\cite{zhuge2024gptswarm}, G-Designer learns task-dependent topologies~\cite{zhang2025gdesigner}, and AFlow searches executable workflows~\cite{zhang2025aflow}.
Graph-of-Agents further selects relevant models and structures their exchanges for each query~\cite{yun2026graphofagents}.
Alongside structure design, another research direction coordinates agents during execution.
Magentic-One uses a central orchestrator to track progress and delegate tasks~\cite{fourney2024magentic};
Evolving Orchestration learns agent-selection policies
~\cite{dang2026multi}, and AOrchestra creates specialised subagents on demand~\cite{ruan2026aorchestra}.
Self-Organizing Agent Teams also retains collaboration experience as reusable organisational strategies
~\cite{pappu2026selforganizing}.
We examine these concerns through duty allocation and information use, then connect agent-proposed decomposition with dependency-guided execution and verification without a central LLM orchestrator.

\begin{figure*}[t!]
  \centering
  \includegraphics[width=\textwidth]{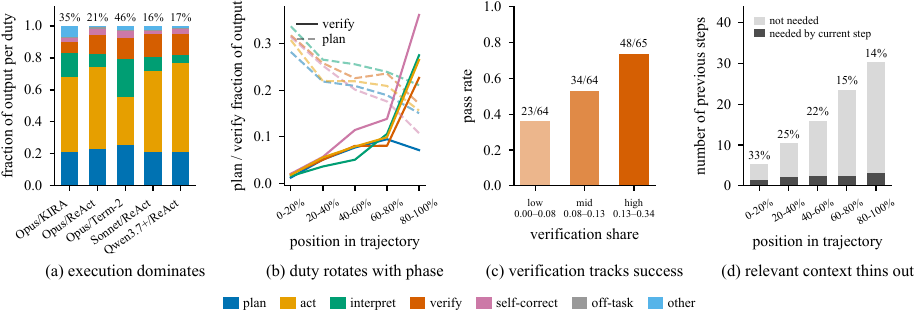}
\caption{(a) Output share by duty; labels show the fraction of adjacent steps with different dominant duties.
(b) Planning (dashed) and verification (solid) output shares across trajectory positions, coloured by configuration.
(c) Pass rates for three equal-sized groups ranked by verification share, pooled across configurations; labels show passing/total trajectories.
(d) Prior steps needed (dark) or not needed (light) by the current duty; labels show the needed fraction.}
  \label{fig:single}
\end{figure*}

\section{Preliminary study }
Long-horizon tasks require agents to execute autonomously with limited human intervention and take on different duties as
execution needs change. We first examine how single agents allocate these duties and
use their shared history to obtain relevant information (Section~\ref{sec:analysis-single}). We then study whether assigning duties to separate agents with their own contexts improves duty organisation and
information use (Section~\ref{sec:analysis-mas}).
\paragraph{Task setting.}
We study long-horizon tasks specified by a natural-language instruction $I$ and an initial environment state $s_0$. An agent uses a policy $\pi$ to select actions based on the instruction and its available context, receiving observations from the environment after each action. A harness manages tool execution, context construction, and termination. In a multi-agent system, it also determines when each agent acts and what information it receives. We evaluate task performance using an external verifier $V$ after execution stops; during execution, agents rely on environment observations and their own checks without access to feedback from $V$. We examine how single-agent and multi-agent systems allocate duties and use information over extended interactions.
\subsection{Auditing long-horizon loops by single agents}
\label{sec:analysis-single}
A long-horizon task requires an agent to perform different duties at different stages of execution. Effective duty allocation is important for detecting errors in time and maintaining progress toward the task goal. In this section, we examine how a single agent manages these duties. Our analysis focuses on two aspects. First, we analyze how verification relates to task success and when it occurs during execution. Second, we examine how information relevant to the current duty is represented in the growing interaction history. Specifically, we first introduce the experimental setup, then present these two analyses and summarize the main findings.

\paragraph{Setup.} We audit 193 trajectories, comprising 4{,}842 agent steps, on a 45-task subset of \texttt{Terminal-Bench~2} whose difficulty composition follows that of the full benchmark (4 easy, 26 medium, 15 hard). We use three harnesses with \texttt{Claude Opus 4.6} (\texttt{Terminus KIRA}, a \texttt{ReAct}-based loop, and \texttt{Terminus 2}), and three base models with the \texttt{ReAct} loop (\texttt{Claude Opus 4.6}, \texttt{Claude Sonnet 4.6}, and
\texttt{Qwen3.7-Plus}). We use another LLM (\texttt{GPT-5.5}) to attribute the content an agent produces at each step, both its stated reasoning and the action it chose, to one of six responsibilities: planning, acting, interpreting an observation, verifying, self-correcting, and working off-task. Content that fits none of them falls into a seventh, unclassified category. Appendix~\ref{app:audit} gives the per-configuration tables and the measurement details.
We summarize the experimental results and observations as follows. 


\textbf{1. The agent spends the least on verification and leaves it to the end, although verification is highly related to task success.}
As shown in Figure~\ref{fig:single}a, verification is the smallest of the main
duties (those above $10\%$ of the output), taking at most $14\%$ of the agent's output. Figure~\ref{fig:single}b shows that it is also the last duty to appear. It is nearly absent early in a trajectory and peaks in the final segment. The agent thus uses verification mainly as a final verify before it stops, rather than interleaving it with execution. Besides, Figure~\ref{fig:single}c also shows that the pass rate rises from $36\%$ in the lowest-verification group to $74\%$ in the highest. No passing trajectory skips verification. These results suggest that verification is closely associated with task success, making its limited and late allocation notable. Verifying more is still not sufficient, since about a quarter of the trajectories in the highest group fail.

\textbf{2. The context mixes all duties, and the proportion relevant to each duty falls as the trajectory grows.}
A single agent keeps the history of all its duties in one context. To measure how much of this context the duty of a given step actually needs, we sample steps from every configuration and ask the judge to mark the previous steps whose content this duty requires. A previous step counts as relevant only if its content is needed, not merely because it performs the same type of duty. As shown in Figure~\ref{fig:single}d, the number of previous steps grows from about $5$ to about $30$ over a trajectory. In contrast, the number of relevant steps grows only from $1.4$ to $3.0$. As a result, the proportion of relevant context falls from $33\%$ to $14\%$. Each duty has to find what it needs among the growing amount of content produced for other duties. We argue that giving each duty its own context would potentially prevent this, and verify whether it is effective in Section~\ref{sec:analysis-mas}.


{\bf Discussion.}
Our study identifies two key limitations of single-agent execution. First, verification often occurs late, despite its strong association with task success, limiting opportunities for correction. Second, records from different duties accumulate in a shared context, reducing the proportion of information relevant to the current duty. This may make it harder for agents to locate the evidence and prior decisions they need. These findings raise a natural question: \textbf{\textit{Can explicitly assigning different duties to separate agents, each with its own context, improve duty allocation and access to relevant information?}}

\begin{figure*}[t!]
  \centering
  \includegraphics[width=\textwidth]{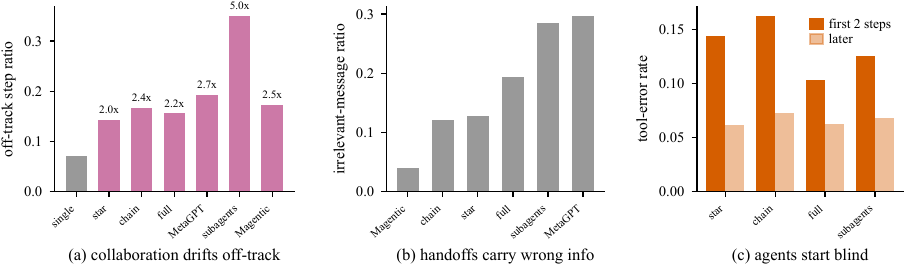}
\caption{
(a) Off-track score of action steps for each system, judged against the original
goal. Labels give the ratio to the single agent.
(b) Fraction of messages between agents that are irrelevant to the receiving agent.
(c) Tool-error rate of delegated agents in their first two steps (dark) and in later
steps (light).}
\label{fig:mas}
\end{figure*}
\subsection{Can multi-agent systems supply what a single agent lacks?}
\label{sec:analysis-mas}
In this section, we answer the above question and explain why assigning duties to separate agents with their own contexts does not by itself resolve the limitations of single-agent execution. Our analysis examines two possible reasons. First, we investigate whether individual agents remain aligned with the overall task objective during collaboration. Second, we examine whether information sharing provides each agent with the background and results needed for its duties. Specifically, we first introduce the experimental setup, then present these two analyses and summarize the main findings.
\paragraph{Setup.}
We use the same 45-task subset of \texttt{Terminal-Bench~2} as in Section~\ref{sec:analysis-single}, and the same policy model \texttt{Claude Sonnet 4.6}. We evaluate six multi-agent paradigms: a MetaGPT-style role pipeline, three static communication topologies (chain, star, and fully connected), a Magentic-One-style orchestrator, and Anthropic-style subagents. With the single-agent baseline, this gives seven configurations. We judge each step with \texttt{GPT-5.5}, using the same responsibility audit as that in Section~\ref{sec:analysis-single}, now applied to each agent separately. We summarize the experimental results and observations as follows:

\textbf{1. Fixed roles fail to keep agents aligned with the overall task objective.} Multi-agent systems rely on collaboration among agents, yet such collaboration does not necessarily ensure that individual agents remain aligned with the overall task objective. To investigate this issue, we employ a judge to evaluate each agent's execution history and determine whether each step contributes to progress toward the overall goal. As shown in Figure~\ref{fig:mas}a, multi-agent systems exhibit off-track behavior $2$–$5$ times more frequently than the single agent. Nearly all off-track steps involve repetitive actions that have already been attempted, indicating that agents become trapped in unproductive cycles. This behavior stems from the rigid organization of existing multi-agent systems: agent roles and execution order are predefined before the task is fully understood, while individual agents operate in isolation without monitoring overall task progress. These systems lack the mechanisms to  dynamically adapt agent responsibilities based on feedback from task execution.


\textbf{2. Agents lack awareness of one another's information needs, leading to ineffective information exchange.} In multi-agent systems, each agent maintains its own context and exchanges messages with other agents to facilitate collaboration. However, we find that $4\%$–$30\%$ of the exchanged messages are irrelevant to their receivers, with systems employing more aggressive message compression exhibiting the highest irrelevance rates (Figure~\ref{fig:mas}b). This inefficiency arises because senders lack awareness of the information needs of their receivers. While message summarization reduces context consumption, it may also discard critical information required by downstream agents. Furthermore, newly delegated agents often lack sufficient background information to perform their assigned tasks effectively. As shown in Figure~\ref{fig:mas}c, their first two execution steps exhibit failure rates up to $2.3$ times higher than those of subsequent steps. Both observations highlight a fundamental limitation: existing systems lack mechanisms for agents to identify and communicate their information needs. Effective multi-agent collaboration therefore requires explicitly modeling information dependencies among agents and enabling information exchange based on these dependencies.

{\bf Discussions.} The above analysis reveals two critical capabilities missing from existing multi-agent systems for long-horizon tasks. The first is \textbf{\textit{goal awareness}}, which requires agents to continuously align their actions with the overall task objective and verify their progress toward that objective throughout execution. Existing systems lack mechanisms to monitor overall task progress, allowing agents to deviate from the goal and repeatedly perform unproductive actions. The second is \textbf{\textit{dynamic collaboration}}, which requires agents to adapt their responsibilities and information exchange based on the evolving needs and dependencies revealed during task execution. Existing systems instead rely on predefined roles and communication structures, preventing agents from performing the right duties at the right time and exchanging the information needed for effective collaboration. Together, goal awareness and dynamic collaboration are essential for multi-agent systems to successfully execute long-horizon tasks.

\begin{figure*}[t]
    \centering
    \includegraphics[width=\textwidth]{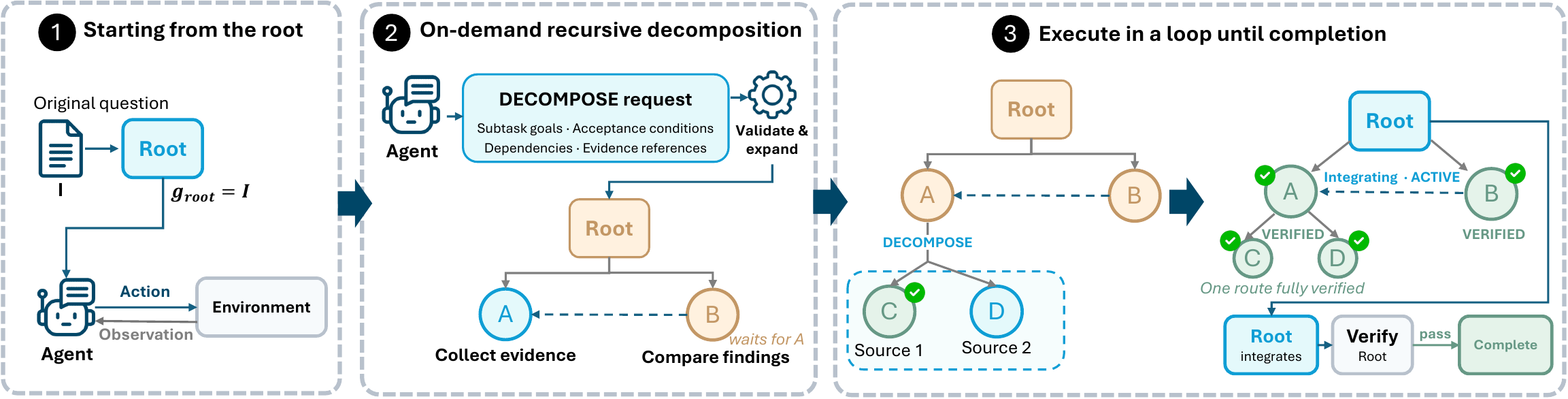}
    \caption{Overview of \ourmethod{}.
(1) Execution starts from a root node representing the original task.
(2) Agents propose subtasks and dependencies as execution needs
emerge; the controller validates these requests before expanding
the graph.
(3) Execution follows dependencies until the root integrates
the results and passes its own verification.}
    \label{fig:grow}
\end{figure*}

\section{Method}
\label{sec:method}
As discussed in the previous section, effective long-horizon collaboration requires goal awareness to keep local tasks aligned with the overall objective, and dynamic collaboration to adapt the division of duties and information sharing to execution needs. To provide these capabilities, we propose \textbf{G}oal-\textbf{A}ware \textbf{D}ynamic Multi-Agent
Collaboration (\ourmethod{}). Our framework builds on a growing execution graph that starts from the original task and expands through agent-proposed subtasks and dependencies. Throughout expansion, each node retains its goal and the duty to satisfy it. We first present this graph and its execution loop in Section~\ref{sec:method-organization}.
We then introduce two supporting mechanisms in Section~\ref{sec:method-support}: dependency-aware context construction supplies relevant information for node execution, while verification and recovery check results and provide feedback for further execution. Figure~\ref{fig:grow} illustrates how decomposition expands the graph and verified results allow execution to continue toward integration and final verification of the original task.
\subsection{The Main Framework: Growing Execution Graph}
\label{sec:method-organization}



    



To support goal awareness and dynamic collaboration, we use a growing execution graph to organise task goals and dependencies.
The graph adapts to emerging execution needs while retaining each node's goal and the duty to satisfy it. In the rest of this subsection, we first define the graph and explain how execution starts from the root. We then introduce agent-proposed decomposition and the controller's validation and graph updates. Finally, we describe the execution loop and how it returns to the original task for integration and completion.

\paragraph{Notations.}
To represent task goals and their dependencies as collaboration develops, we maintain an execution graph
$\mathcal{G}_t=(\mathcal{N}_t,\mathcal{E}_t)$, where $t$ indexes scheduling steps. Each node $v$ represents an assigned task and stores its specific goal $g_v$, acceptance conditions $K_v$ that determine whether the goal has been achieved, and an execution status $\sigma_v$. Edges connect a task to its subtasks, or to another task whose verified result it requires.

\textbf{Step 1. Starting from the root.}
Initially, $\mathcal{N}_0=\{v_{\mathrm{root}}\}$,
$\mathcal{E}_0=\varnothing$, and $g_{v_{\mathrm{root}}}=I$.
We include full satisfaction of $I$ as the root's acceptance condition and then launch an agent whose initial context contains the instruction, this completion requirement, and the root's status. The graph has not yet expanded, so the root has no dependency results to draw on. Its execution history starts empty and accumulates observations as the agent interacts with the environment. These records guide subsequent actions without requiring a complete workflow in advance. The agent starts by interacting directly with the environment and uses the resulting observations to guide subsequent actions, without constructing a complete workflow in advance.

\textbf{Step 2. On-demand recursive decomposition.} As agents execute their assigned tasks, new observations may reveal needs for additional subtasks. To adapt the division of duties to these needs, we allow each executing agent to request decomposition. We describe when the agent makes this decision, what it submits, and how the controller uses the request to update the graph.

\begin{itemize}[leftmargin=1.5em]
    \item \textit{When to decompose.}
We use on-demand decomposition to adapt the division of duties to needs that emerge during execution.
The agent executing each node decides whether to introduce subtasks, without relying on a central LLM orchestrator or a predefined communication topology.
We prompt the agent executing node $v$ to express its next
decision as a structured action $u_{t,v}$.
The core action types are
\begin{equation}
\mathcal{U}_{\mathrm{core}} =
\{\texttt{RUN},\texttt{DECOMPOSE},\texttt{SEARCH},\texttt{VERIFY}\}.
\label{eq:action-space}
\end{equation}
These actions respectively execute an environment command,
request decomposition, retrieve shared artifacts, and request
verification.
When $\operatorname{type}(u_{t,v})=\texttt{DECOMPOSE}$,
the agent prepares a decomposition request as follows.

\item \textit{Submitting decomposition requests.}
The agent submits a request
\begin{equation}
d_v =
\left(
\{(\ell_i,g_i,K_i,P_i)\}_{i=1}^{m},J_v
\right),
\label{eq:decomposition-request}
\end{equation}
where $\ell_i$ identifies a proposed subtask and $g_i$ is its goal.
The acceptance conditions $K_i$ specify its completion requirements,
which we assess through the verification scheme in
Section~\ref{sec:method-verification}.
The dependency set $P_i$ lists the subtasks or existing nodes
whose verified results it requires.
These dependencies guide the supply of results to its execution
context through the scheme in Section~\ref{sec:method-information}.
The set $J_v$ references previous execution steps recorded at $v$,
grounding the decomposition in execution evidence.
The agent submits $d_v$ to the controller for validation.
\item {\textit{How to update the graph.}}
Once an agent submits a decomposition request $d_v$, we need to determine whether it can extend the graph and how execution should proceed after the update. We introduce a lightweight, rule-based harness controller $\mathcal{H}$ to validate the request and apply graph updates without additional model inference. It checks $d_v$ against the current graph, the node-specific execution history $h_t^v$, and the remaining resource budgets $B_t$:
\begin{equation}
(b_v,f_v)
=
\operatorname{Check}_{\mathcal{H}}
(d_v,\mathcal{G}_t,h_t^v,B_t),
\label{eq:decomposition-check}
\end{equation}
where $b_v\in\{0,1\}$ indicates whether the request is accepted and $f_v$ provides feedback. The checks require at least one valid reference to an execution step of $v$, valid subtask and dependency specifications, compliance with resource limits, and an acyclic resulting graph. If $b_v=0$, the graph remains unchanged and the agent receives $f_v$ to revise its next action.
Otherwise, $\mathcal{H}$ creates or reuses compatible child nodes $\mathcal{C}_v$, resolves the references in $P_i$ into dependency edges $\mathcal{D}_v$, and updates the graph:
\begin{align}
\mathcal{N}_{t+1}
&= \mathcal{N}_t \cup \mathcal{C}_v, \nonumber\\
\mathcal{E}_{t+1}
&= \mathcal{E}_t
   \cup \{(v,c):c\in\mathcal{C}_v\}
   \cup \mathcal{D}_v.
\label{eq:graph-expansion}
\end{align}
Here, $(x,y)\in\mathcal{D}_v$ means that $x$ requires the verified result of $y$. The child set forms a new route $\rho_{v,k}=\mathcal{C}_v$,
where $k$ indexes alternative approaches to achieving $g_v$. Node $v$ waits while the route remains unresolved.
To maintain goal awareness during graph expansion, we keep
$g_v$ and its full-goal acceptance condition unchanged, with the root goal remaining $I$.
Even when a route completes, node $v$ must still pass verification against its original goal $g_v$.

\end{itemize}

\textbf{Step 3. Execute in a loop until completion.}
After updating the graph, $\mathcal{H}$ selects nodes for subsequent execution. The controller tracks each node's status through four core states:
\begin{equation}
\sigma_v \in
\{\texttt{READY},\texttt{ACTIVE},\texttt{WAITING},\texttt{VERIFIED}\}.
\label{eq:node-states}
\end{equation}
These indicate that a node can execute, is currently executing, awaits required results, or has passed verification, respectively. The controller updates these states from dependency and route outcomes and identifies the nodes still needed for the root task, denoted by $\mathcal{N}^{\mathrm{needed}}_t$. It constructs the execution frontier
\begin{equation}
\mathcal{F}_t =
\left\{
v\in\mathcal{N}^{\mathrm{needed}}_t
\mid \sigma_v=\texttt{READY}
\right\}.
\label{eq:ready-frontier}
\end{equation}
For each selected node, $\mathcal{H}$ launches an agent with a local context constructed through our dependency-guided scheme in Section~\ref{sec:method-information}. This context supplies the recorded observations and relevant results needed to continue the assigned task. The agent executes its task, requests decomposition, or requests verification as described in Section~\ref{sec:method-verification}. As execution changes the graph and node states, $\mathcal{H}$
recomputes the frontier and repeats this process.

A node becomes ready once its required dependencies are verified. When a route completes and the other dependencies of node $v$ are satisfied, $v$ resumes to integrate the results and pass its own verification. If all its routes fail, $v$ can resume to attempt another approach. The system declares task completion only when the root passes verification against the original instruction $I$ and all acceptance conditions in $K_{v_{\mathrm{root}}}$. This can follow direct execution or subtask integration. Execution also stops when the budget runs out or no nodes can execute and no agents remain active, but these cases do not establish task completion.

\subsection{Supporting Mechanisms: Context Construction and Verification}
\label{sec:method-support}
To support execution of the growing graph, we need to provide agents with relevant information and assess whether their results satisfy the assigned goals. We introduce two supporting mechanisms: dependency-aware context construction supplies the inputs for each node's execution, while verification and recovery check its results and provide feedback for further execution. We describe these mechanisms in Sections~\ref{sec:method-information} and~\ref{sec:method-verification}, respectively.

\subsubsection{Dependency-Aware Context Construction}
\label{sec:method-information}

As mentioned in Section~\ref{sec:method-organization}, the execution graph determines which nodes can execute next. The agents launched for these nodes also need relevant background and results to continue execution.
Our analysis in Section~\ref{sec:analysis-mas} shows that agents can lack this information or receive messages unrelated to their needs. We address this gap through two complementary mechanisms: dependency-based sharing provides verified results from declared dependencies, while dependency-free retrieval supplies additional information without requiring explicit dependency links.

\textbf{Context formulation.} We maintain a shared collection of results $\mathcal{R}_t$, where the agent executing each node can contribute findings with references to supporting execution steps. To share these findings without copying the contributing agent's full history, the controller selects relevant results and combines them with the node's task information and execution records.
For node $v$, this gives the context
\begin{equation}
c_{t,v} =
\left(
I,g_v,K_v,\mathcal{L}_t(v),
\mathcal{R}^{\mathrm{dep}}_t(v),
\mathcal{R}^{\mathrm{ret}}_t(v)
\right),
\label{eq:local-context}
\end{equation}
where $\mathcal{L}_t(v)$ contains recent execution records, the node's own results, and information about the task that introduced $v$, its dependencies, and its routes. These records support continued execution, while $I$, $g_v$, and $K_v$ retain the overall goal and current task requirements. Given $c_{t,v}$, the agent generates its next structured action
as $u_{t,v}\sim\pi(\cdot\mid c_{t,v})$, following
Eq.~\eqref{eq:action-space}.

\textbf{Dependency-based sharing.}
The controller uses node $v$'s declared dependencies to identify the results it requires.
It directly includes verified results from these dependencies in $\mathcal{R}^{\mathrm{dep}}_t(v)$. This connects the dependencies proposed during decomposition to the information available during subsequent execution.

\textbf{Dependency-free retrieval.}
Declared dependencies may not cover all the information needed for the next action.
The controller supplements these inputs with relevant results retrieved from $\mathcal{R}_t$, denoted by $\mathcal{R}^{\mathrm{ret}}_t(v)$.
Retrieval does not require an explicit dependency link to $v$. When the agent needs further information, it can autonomously output a \texttt{SEARCH} action with a query. The controller includes the retrieved results in the next context for $v$.

\subsubsection{Verification and Recovery}
\label{sec:method-verification}
As mentioned in Section~\ref{sec:method-organization}, verified node results allow subsequent tasks to proceed and enable
result integration. This requires checking whether each node has satisfied its goal when its agent requests completion. Our analysis in Section~\ref{sec:analysis-single} shows that
agents often defer verification until late in execution, leaving little opportunity to repair errors. We address this issue through two stages:
node verification combines executable checks with an independent semantic review to assess completion; result acceptance and repair uses this assessment to accept the result or return feedback for further execution.

\textbf{Node Verification.}
When the agent executing node $v$ outputs \texttt{VERIFY}, the controller runs the executable checks in $K_v$ and appends their observations to $h_t^v$. Let $m_v=1$ indicate that all these checks pass. The controller then invokes a semantic verifier with a fresh context:
\begin{equation}
q_v =
\begin{cases}
\operatorname{Audit}_{\pi}(v), & m_v=1,\\
0, & \text{otherwise}.
\end{cases}
\label{eq:node-verification}
\end{equation}
Here, $\operatorname{Audit}_{\pi}(v)\in\{0,1\}$ assesses whether the execution evidence and verified dependency results support $g_v$ and every acceptance condition in $K_v$. The verifier must address each condition and cite supporting evidence. When no executable check applies, we set $m_v=1$ and assess the recorded evidence directly. This internal review differs from the external task verifier $V$, which evaluates the final environment after execution stops.

\textbf{Result Acceptance and Repair.}
If $q_v=1$, the controller sets $\sigma_v=\texttt{VERIFIED}$ and adds an evidence-linked result to $\mathcal{R}_t$ for subsequent dependency contexts. Otherwise, it returns the failed check or review feedback to node $v$ for continued execution and repair. A rejected verification does not by itself fail the route. Both outcomes return to the execution loop in Section~\ref{sec:method-organization}. The pseudo-code algorithm of
\ourmethod{} is provided in Appendix ~\ref{app:alg}.
\begin{table*}[t]
\centering
\small
\caption{Task resolution rate (\%) on Terminal-Bench 2.1 and
SWE-Bench Pro Hard-51 and all methods with same execution
budget.
$^{\dagger}$LongHorizon-Harness drives Claude code is
incompatible with SWE-Bench Pro's \texttt{no-network} agent policy.}
\setlength{\tabcolsep}{10pt}
\renewcommand{\arraystretch}{1.0}
\begin{tabular}{llcccc}
\toprule
 & & \multicolumn{2}{c}{\textbf{Terminal-Bench 2.1}} & \multicolumn{2}{c}{\textbf{SWE-Bench Pro (Hard-51)}} \\
\cmidrule(lr){3-4}\cmidrule(lr){5-6}
\textbf{Category} & \textbf{Method} & Sonnet\,4.6 & Qwen3.7-Plus & Sonnet\,4.6 & Qwen3.7-Plus \\
\midrule
Single agent          & Single (React loop)             & 57.3 & 56.2 & 27.5 & 31.4 \\
& Single (Terminus-KIRA)           & 41.6 & 41.6 &  49.0  & 31.4 \\
& Single (Terminus-2)             & 52.8 & 55.0 & 64.7 &  27.5\\
\midrule
\multirow{5}{*}{\shortstack[l]{Static MAS\\(fixed topology)}}
                      & Chain               & 58.4 & 51.7 & 39.2 & 25.5 \\
                      & Star                & 56.2 & 52.8 & 43.1 & 15.7 \\
                      & Full                & 56.2 & 49.4 & 39.2 & 23.5 \\
                      & MetaGPT             & 46.1 & 51.7 & 31.4 & 29.4 \\
                      & Mixture-of-Agents   & 40.4 & 51.7 & 13.7 & 25.5 \\
\midrule
\multirow{4}{*}{\shortstack[l]{Dynamic MAS\\(adaptive)}}
                      & Subagents           & 52.8 & 51.7 & 17.6 & 27.5 \\
                      & Magentic-One        & 36.0 & 48.3 & 27.5 & 25.5 \\
                      & LongHorizon-Harness (React loop) & 57.3 & 56.2 & 11.8 & 17.6 \\
                      & LongHorizon-Harness (Claude code) & 65.2 & 57.3 & --$^{\dagger}$ & --$^{\dagger}$ \\
\midrule
Ours                  & \textbf{\ourmethod{}} & \textbf{71.9} & \textbf{64.0} & \textbf{66.7} & \textbf{33.3} \\
\bottomrule
\end{tabular}
\label{tab:main}
\end{table*}
\section{Experiments}
In this section, we first describe the experimental setup in Section~\ref{sec:experimental-setup} and present the main results in Section~\ref{sec:main-results}. We then conduct ablation studies in Section~\ref{sec:ablation}. Next, we evaluate how our approach addresses the two limitations of MAS in Section~\ref{sec:analysis-depth}. Finally, we analyze its computational efficiency in Section~\ref{sec:efficiency}.

\subsection{Experimental Setup}
\label{sec:experimental-setup}
\paragraph{Datasets.}
We evaluate \ourmethod{} on two challenging benchmarks, namely
\texttt{Terminal-Bench~2.1}~\cite{merrill2026terminal}
and the Hard-51 subset of the \texttt{SWE-Bench Pro~V2}
~\cite{deng2025swe}. Specifically, \texttt{Terminal-Bench~2.1} contains 89 tasks covering diverse terminal operations. Hard-51 contains 51 challenging repository-level software engineering tasks. We report task resolution rates using the benchmark verifiers. For \texttt{SWE-Bench Pro}, we evaluate generated patches in fresh environments following the official protocol.

\paragraph{Baselines.}
Our single-agent baselines include the following three methods:
\texttt{ReAct} based loop~\cite{yao2022react}, \texttt{Terminus-KIRA},
\texttt{Terminus-2}. For static multi-agent collaboration, we compare Chain, Star, Full~\cite{zhang2025gdesigner}, MetaGPT~\cite{hong2024metagpt}, and Mixture-of-Agents~\cite{wang2025mixture}. For dynamic collaboration, we include  Subagents, Magentic-One~\cite{fourney2024magentic}, and LongHorizon-Harness~\cite{ma2026longhorizon} with ReAct and Claude Code backends.

\paragraph{Implementation details.}
We use Claude Sonnet~4.6 and Qwen3.7-Plus as the base models. We run tasks in containerised environments, with Docker as the default backend. Our method uses the same base model for node execution and semantic verification.
We allow up to three concurrent workers per task and include the seven most recent node execution records in each local context. We retrieve up to six shared results per query.
Our method permits up to 1,000 tool calls and 2,000 model calls per task, without an aggregate token limit. The wall-clock limit is 18,000 seconds on Terminal-Bench; the available Hard-51 runs use 3,000 or 4,500 seconds. Appendix~\ref{app:implementation} provides the detailed settings.

\subsection{Main Results}
\label{sec:main-results}In this section, we compare \ourmethod{} with three single-agent harnesses
and nine multi-agent systems. Table~\ref{tab:main} reports the resolution rates. The main findings are as follows.

\textbf{\ourmethod{} consistently outperforms the single-agent harnesses.}
As shown in Table~\ref{tab:main}, \ourmethod{} achieves the highest
resolution rate across all benchmarks and base models.
Compared with the strongest single-agent harness in each setting,
it improves resolution rates on \texttt{Terminal-Bench~2.1} by
14.6 and 7.8 percentage points under \texttt{Sonnet~4.6} and
\texttt{Qwen3.7-Plus}, respectively.
The corresponding gains on Hard-51 of \texttt{SWE-Bench Pro}  are
$2.0\%$ and $1.9\%$.
The improvements are larger on Terminal-Bench, while the strongest single-agent harness remains competitive on Hard-51. We suggest that these improvements result from separating local task execution from independent verification, so checking does not rely solely on the executing agent's own duty allocation. Node-specific contexts also let agents continue from relevant records without recovering their task state from the mixed history of all duties.

\textbf{\ourmethod{} achieves higher resolution rates than existing
multi-agent designs.}
As shown in Table~\ref{tab:main}, \ourmethod{} exceeds the strongest
MAS baseline by $6.7$\% and $6.7\%$
on \texttt{Terminal-Bench~2.1}, and by $23.6\%$ and $3.9\%$ 
on Hard-51, under two base model, respectively.
The baseline results also show that multi-agent designs do not consistently outperform single-agent harnesses, and dynamic designs do not always improve over fixed topologies. These observations support our preliminary finding that separating duties across agents alone does not ensure effective collaboration. We suggest that these improvements result from adapting the division of duties to execution needs while retaining the original goal and each node's acceptance conditions. Explicit dependencies also connect task scheduling with the results each agent receives, addressing the missing background and irrelevant information exchange observed in existing MAS.

In summary, we suggest that these gains result from on-demand decomposition, which lets agents focus on subtasks revealed during execution, and dependency-aware context construction, which provides the prior results needed to continue without repeating earlier investigations.

\subsection{Ablation Studies}
\label{sec:ablation}
We ablate goal awareness by removing goal-related tracking mechanisms, and dynamic collaboration by disabling helper requests, inter-agent dependency requests, and decomposition. We then isolate the contribution of decomposition on tasks where
\ourmethod{} activates it. The main findings are as follows.

\begin{table}[h]
  \centering
  \small

  \caption{Ablation of \ourmethod{}. Pass rate (\%) in brackets}

  \setlength{\tabcolsep}{3pt}
  \label{tab:ablation}
  \begin{tabular}{lcccc}
    \toprule
    & \multicolumn{2}{c}{\texttt{Sonnet~4.6}} & \multicolumn{2}{c}{\texttt{Qwen3.7-Plus}} \\
    \cmidrule(lr){2-3} \cmidrule(lr){4-5}
    Method & \texttt{TB2.1} & \texttt{Hard51} & \texttt{TB2.1} & \texttt{Hard51} \\
    \midrule
    \ourmethod{} & \textbf{71.9} (64) & \textbf{66.7} (34) & \textbf{64.0} (57) & \textbf{33.3} (17) \\
   w/o goal awareness & 69.7 (62) & 58.8 (30) & 60.7 (54) & 27.5 (14) \\
     w/o dynamic collaboration & 67.4 (60) & 52.9 (27) & 61.8 (55) & 23.5 (12) \\
    \bottomrule
  \end{tabular}
\end{table}
\begin{figure*}[t]
  \centering
  \includegraphics[width=\textwidth]{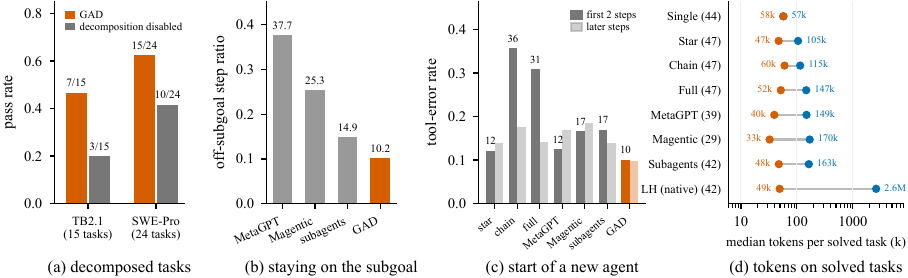}
 \caption{
(a) Pass rate on the decomposed tasks, with and without decomposition.
(b) Fraction of steps that leave the assigned subgoal.
(c) Tool-error rate of a new agent in its first two steps and later.
(d) Each row compares \ourmethod{} (orange) with one baseline (blue) on the
tasks both solve. Dots show median tokens per task; baseline counts indicate task numbers.}
  \label{fig:depth}
\end{figure*}
\textbf{Both goal awareness and dynamic collaboration contribute
to performance.}
As shown in Table~\ref{tab:ablation}, removing either component lowers the resolution rate in all four settings. With \texttt{Sonnet~4.6} on \texttt{SWE-Bench Pro}, the resolution rate falls from $66.7\%$ to $58.8\%$ without goal awareness and to $52.9\%$ without dynamic collaboration. With \texttt{Qwen3.7-Plus} on \texttt{Terminal-Bench~2.1}, removing goal awareness leads to a larger drop than removing
dynamic collaboration, with resolution rates of $60.7\%$ and $61.8\%$, respectively.
These results suggest that the two components contribute in different ways. Goal awareness applies throughout execution, whereas decomposition changes the task structure only when an agent requests it. \texttt{Qwen3.7-Plus} decomposes only 2 of the 89 \texttt{Terminal-Bench~2.1} tasks, which limits the contribution of decomposition in this setting.

\textbf{On-demand decomposition improves task success.}
We select the tasks that \ourmethod{} decomposes in its full configuration and rerun the same tasks with decomposition disabled.
Each task must then proceed through the root node without creating subtasks. As shown in Figure~\ref{fig:depth}a, the number of solved tasks falls from 7 to 3 among the 15 selected
\texttt{Terminal-Bench~2.1} tasks, and from 15 to 10 among the 24 selected \texttt{SWE-Bench Pro} tasks. These results suggest that decomposition contributes to success on tasks where \ourmethod{} chooses to introduce subtasks. 

\subsection{Improvement on the two limitations in MAS}
\label{sec:analysis-depth}
To examine whether \ourmethod{} provides the \textbf{\textit{goal awareness}} and \textbf{\textit{dynamic collaboration}} that existing multi-agent systems lack, we audited its trajectories with the procedure of Section~\ref{sec:analysis-mas}. \ourmethod{} decomposes seven of the 45 \texttt{Terminal-Bench~2.1} tasks. Only these trajectories involve more than one agent, so the analysis uses these seven tasks. All systems run on \texttt{Sonnet~4.6}.

\textbf{Agents in \ourmethod{} stay on the subgoals they are assigned.}
As shown in Figure~\ref{fig:depth}b, when each execution step is judged against the subgoal assigned to the agent that took it, only $10.2\%$ of the steps of \ourmethod{} leave their subgoal. The rate is $14.9\%$ for {subagents} and $37.7\%$ for {MetaGPT}. We suggest that a subgoal formed from execution gives each agent a clear target, because it carries its own acceptance criteria. Each agent thus keeps its work tied to the goal.


\textbf{Agents in \ourmethod{} start working as soon as they are assigned.}
As shown in Figure~\ref{fig:depth}c, $10.1\%$ of the first two steps of a newly assigned agent in \ourmethod{} end in a tool error. Its later steps fail at a similar rate of $9.8\%$. In {chain}, the first two steps fail in $35.7\%$ of cases, twice its later rate of $17.6\%$. We suggest that \ourmethod{} gives each agent the verified results of the subgoals it depends on. The agent thus starts from what is already known and does not need to explore the environment first.
\subsection{Efficiency Analysis}
\label{sec:efficiency}
We measure cost by the tokens a system consumes on a task. The count adds the input and output tokens of every LLM call made by every agent in the system, including the calls \ourmethod{} makes for verification. The main findings are as follows.
\textbf{\ourmethod{} solves the same tasks at lower cost.}
We compare each baseline with \ourmethod{} on the tasks both systems solve, and
report the median token usage.
As shown in Figure~\ref{fig:depth}d, on \texttt{Terminal-Bench 2.1} with \texttt{Sonnet 4.6}, \ourmethod{} needs about 50k tokens per task. Multi-agent systems need 105--170k on the same tasks, and the long-horizon harness needs 2.6M. 
We suggest that the cost of existing multi-agent systems comes from coordination that they pay regardless of the task. \ourmethod{} adds structure only when the goal requires it. On simple tasks it runs like a single agent, so its cost matches the single agent and stays below systems that coordinate on every task. Its accuracy gain does not come from spending more computation.

\section{Conclusion}
Long-horizon tasks require LLM agents to manage multiple duties throughout repeated interactions with the environment. Our analysis shows that verification has the strongest association with task success but often comes late, while the proportion of duty-relevant information decreases as the context grows. Existing multi-agent systems can lose track of the overall goal, and context separation alone does not ensure effective information sharing. We propose \ourmethod{} to maintain goal awareness and organize collaboration as execution reveals new needs. It retains the original instruction and assigned task goals and acceptance conditions in each agent's context. A growing execution graph guides scheduling and information sharing as agents divide tasks and specify dependencies during execution. Verification and recovery check both subtask results and full task completion, providing feedback for repair. Experiments with Claude Sonnet~4.6 and Qwen3.7-Plus on Terminal-Bench~2.1 and SWE-bench Pro Hard-51 show gains of up to 6.7 percentage points over the strongest baseline.
\bibliographystyle{ACM-Reference-Format}
\bibliography{references}
\clearpage
\appendix
\section{Details of the Audit}
\label{app:audit}

\paragraph{Configurations.}
We audit five configurations on the 45-task subset of \texttt{Terminal-Bench~2}
(4 easy, 26 medium and 15 hard tasks). The first three vary the harness with
\texttt{Opus~4.6} fixed. The last three vary the model with ReAct fixed.
Trajectories whose verifier did not run are excluded, since they are neither
passes nor failures. Table~\ref{tab:app-single-cells} lists the trajectories
that remain.

\begin{table}[h]
  \centering
  \small
  \caption{Audited single-agent configurations.}
  \label{tab:app-single-cells}
  \begin{tabular}{llrrr}
    \toprule
    Model & Harness & Trajectories & Steps & Pass rate (\%) \\
    \midrule
    \texttt{Opus~4.6}     & \texttt{Terminus-KIRA} & 43 & 1544 & 62.8 \\
    \texttt{Opus~4.6}     & ReAct                  & 44 & 1262 & 43.2 \\
    \texttt{Opus~4.6}     & \texttt{Terminus-2}    & 29 &  418 & 41.4 \\
    \texttt{Sonnet~4.6}   & ReAct                  & 32 &  456 & 62.5 \\
    \texttt{Qwen3.7-Plus} & ReAct                  & 45 & 1162 & 60.0 \\
    \bottomrule
  \end{tabular}
\end{table}

\paragraph{Duty attribution.}
A judge splits the content each step generates into the seven duties of
Table~\ref{tab:app-duties}, as fractions that sum to one. The judge sees the task
instruction and, for each step, the agent's reasoning, its command and the
resulting observation. Our own step labels are not shown, since they would reveal
the answer. Table~\ref{tab:app-judge} gives the settings.

\begin{table}[h]
  \centering
  \small
  \caption{Duties used in the audit.}
  \label{tab:app-duties}
  \begin{tabular}{lp{0.68\linewidth}}
    \toprule
    Duty & Definition \\
    \midrule
    plan         & deciding what to do next, ordering subgoals, tracking the objective \\
    act          & issuing the command, or writing the code it contains \\
    interpret    & reading and making sense of the previous observation \\
    verify       & checking whether the result meets the acceptance criteria \\
    self-correct & diagnosing an earlier mistake and recovering from it \\
    off-task     & content unrelated to the task, or repeating covered ground \\
    other        & legitimate content that fits none of the above \\
    \bottomrule
  \end{tabular}
\end{table}

\begin{table}[h]
  \centering
  \footnotesize
  \setlength{\tabcolsep}{4pt}
  \caption{Judge and metric settings.}
  \label{tab:app-judge}
  \begin{tabular}{lp{0.7\linewidth}}
    \toprule
    Setting & Value \\
    \midrule
    Judge model & \texttt{GPT-5.5}, temperature 1.0 \\
    Steps per call & 5 \\
    Text per step & reasoning 900, command 600, observation 400 characters \\
    Task instruction & 1500 characters \\
    Dominant duty & duty with the largest fraction \\
    Duty-switch rate & share of adjacent steps with different dominant duties \\
    Position & five equal trajectory segments \\
    Verification groups & thirds by verification share, pooled \\
    Confidence intervals & 2000 bootstrap draws over trajectories \\
    \bottomrule
  \end{tabular}
\end{table}
\paragraph{Context relevance.}
We sample 30 steps from each configuration, 6 from each of the five trajectory
segments. A sampled step lies in a trajectory of at least six steps and has at
least three steps before it. The judge sees the task, the current step and every
earlier step with its observation. It lists the earlier steps whose content the
current step needs. A step on the same topic or of the same duty is not listed
unless its content is needed. The reported share is the number of listed steps
over the number of earlier steps. Earlier observations are cut to 8000
characters, the length the harnesses keep in context, and the rendered history to
60000 characters.
\section{Pseudo-code for \ourmethod{}}
\label{app:alg}
Algorithm~\ref{alg:method} presents the core workflow of
\ourmethod{} in sequential form.
Lines~1--2 initialize the graph, shared results, environment,
and root task.
At each iteration, Lines~4--9 update node states, check stopping
conditions, select a ready node, and generate an action from
its local context.
Lines~10--18 handle environment execution, decomposition,
and retrieval.
An accepted decomposition expands the graph, while a rejected
request returns feedback.
Lines~19--22 check node completion and either publish a verified
result or retain feedback for repair.
Line~24 returns a node to \texttt{READY} if it remains
\texttt{ACTIVE}.
The loop stops when the root passes verification, the budget
runs out, or no ready nodes remain.
Only root verification establishes task completion.
In concurrent execution, the controller waits for running
agents before treating an empty frontier as a stopping condition.
\begin{algorithm}[t]
\caption{Workflow of \ourmethod}
\label{alg:method}
\begin{algorithmic}[1]
\Require Instruction $I$, environment $s_0$, policy $\pi$, budget $B$
\Ensure Final environment state and root completion status
\State Initialize $\mathcal{G}=(\{v_{\mathrm{root}}\},\varnothing)$,
       $\mathcal{R}\gets\varnothing$, and $s\gets s_0$
\State Set root goal to $I$, acceptance to full satisfaction of $I$,
       and status to \texttt{READY}
\While{\textbf{true}}
    \State Update node states from dependency and route outcomes,
           following Section~\ref{sec:method-organization}
    \State \textbf{break} if root is \texttt{VERIFIED} or budget is exhausted
    \State Compute $\mathcal{F}$ by Eq.~\eqref{eq:ready-frontier};
           \textbf{break} if $\mathcal{F}=\varnothing$
    \State Select $v\in\mathcal{F}$; launch an agent and set
           $\sigma_v\gets\texttt{ACTIVE}$
    \State Construct $c_{t,v}$ by Eq.~\eqref{eq:local-context}
    \State Sample $u_{t,v}\sim\pi(\cdot\mid c_{t,v})$;
           register evidence-backed findings in $\mathcal{R}$
    \If{$u_{t,v}$ requests \texttt{RUN}}
        \State Execute its command; update $s$ and append observations to $h^v$
    \ElsIf{$u_{t,v}$ requests \texttt{DECOMPOSE}}
        \State Obtain $(b_v,f_v)$ by checking $d_v$
               using Eq.~\eqref{eq:decomposition-check}
        \State If $b_v=0$, retain $f_v$, set $v$ \texttt{READY},
               and \textbf{continue}
        \State Create or reuse $\mathcal{C}_v$ and resolve
               dependency edges $\mathcal{D}_v$
        \State Expand $\mathcal{G}$ by Eq.~\eqref{eq:graph-expansion};
               register $\rho_{v,k}=\mathcal{C}_v$; set $v$ \texttt{WAITING}
    \ElsIf{$u_{t,v}$ requests \texttt{SEARCH}}
        \State Retrieve results using its query for the next context;
               set $v$ \texttt{READY}
    \ElsIf{$u_{t,v}$ requests \texttt{VERIFY}}
        \State Compute $q_v$ by Eq.~\eqref{eq:node-verification}
        \State \textbf{if} $q_v=1$, set $v$ \texttt{VERIFIED}
               and publish its verified result in $\mathcal{R}$
        \State \textbf{else} retain verification feedback
               and set $v$ \texttt{READY}
    \EndIf
    \State Return $v$ to \texttt{READY} if it remains \texttt{ACTIVE}
\EndWhile
\end{algorithmic}
\end{algorithm}

\section{Implementation Details}
\label{app:implementation}

We use Claude Sonnet~4.6 and Qwen3.7-Plus as the base models, with the same
model for node execution and semantic verification within each setting.
Tasks run in containerized environments, with Docker as the default backend.
Table~\ref{tab:implementation-hparams} summarizes the reference Growing-DAG
V1.1 configuration. All methods use the same per-task execution budget
within each evaluation setting. For Growing-DAG, tool and model-call budgets
are shared by all workers within a task, and verification commands count
toward the tool-call budget. No aggregate token limit is imposed.
The working deadline is 18,000 seconds for Terminal-Bench and 3,000 or
4,500 seconds for Hard-51, depending on the evaluation setting.
Additional harness time for cleanup and patch capture does not extend
the model's working deadline.

Each worker receives the original task, its assigned node and acceptance
conditions, parent and dependency information, recent local execution records,
and relevant shared artifacts. Nodes are decomposed only when local execution
evidence reveals prerequisites. Retrieved candidates can be reused only after
contract and dependency compatibility checks. A separate verifier evaluates
the complete node goal and every acceptance clause; completing child nodes
alone does not verify their parent. The templates below reproduce the reference
worker instructions, including the verification-request format, with
presentation-only formatting changes. Braced input fields are populated by
the runtime. The worker prompt's \texttt{\{max\_children\}} and
\texttt{\{max\_routes\_per\_node\}} take the configured values in
Table~\ref{tab:implementation-hparams}.

\begin{table}[t]
\centering
\small
\caption{Reference \ourmethod{} hyperparameters and per-task budgets.
Output allowances are initial per-response limits; response recovery may
increase them up to 32,768 tokens. Saved run overrides take precedence.}
\label{tab:implementation-hparams}
\begin{tabularx}{\linewidth}{@{}Xr@{}}
\toprule
Hyperparameter & Value \\
\midrule
Concurrent workers & 3 \\
Recent local records (worker / verifier) & 7 / 10 \\
Retrieved artifacts per query & 6 \\
Maximum nodes / depth & 64 / 8 \\
Maximum children per route & 4 \\
Maximum routes per node & 3 \\
Reuse candidate similarity threshold & 0.72 \\
Temperature (worker / verifier) & 0.1 / 0.0 \\
Initial output tokens (worker / verifier) & 8,192 / 4,096 \\
Tool / model calls per task & 1,000 / 2,000 \\
Aggregate token budget & Unlimited \\
Working deadline: Terminal-Bench (s) & 18,000 \\
Working deadline: Hard-51 (s) & 3,000 or 4,500 \\
\bottomrule
\end{tabularx}
\end{table}

\begin{tcolorbox}[
  enhanced, breakable,
  colback=white, colframe=black!40,
  boxrule=0.5pt, arc=1mm
]
\subsection*{Role and Coordination Rules}

\noindent You are a disposable LOCAL WORKER in a long-horizon terminal task system.
There is NO Planner. You own exactly one obligation node in a lazily growing execution DAG.
Do not produce a global plan. Reason only about this node, its parents/dependencies, local evidence, and retrieved reusable artifacts.

\noindent The DAG is a hard coordination contract:
\begin{enumerate}[leftmargin=*, label=\arabic*.]
    \item Work only on the assigned node.
    \item A node may be solved directly; decomposition is NOT mandatory.
    \item DECOMPOSE only after concrete local execution evidence reveals independent prerequisites/blockers. "The task is broad" is NOT sufficient.
    \item Every DECOMPOSE must cite one or more prior local evidence step IDs; the runtime rejects uncited graph growth.
    \item Decomposition creates ONE alternative AND-route: every child in that route must be VERIFIED before the parent can be retried.
    \item The parent NEVER becomes verified merely because children finish. It must later pass its own acceptance check.
    \item New children must be narrower/more local/more verifiable than the parent. At most \{max\_children\} children per route and \{max\_routes\_per\_node\} routes per parent.
    \item Prefer SEARCH/reuse over duplicate work.
    \item Never access hidden benchmark solutions, hidden verifier state, credentials, or host data.
\end{enumerate}

\noindent Collaboration is artifact-mediated, not conversational. You may publish claims supported by PRIOR local step IDs. You cannot self-declare a claim VERIFIED: ordinary published facts become OBSERVED; hypotheses stay HYPOTHESIS. Only runtime verification creates VERIFIED artifacts.

\end{tcolorbox}

\begin{tcolorbox}[
  enhanced, breakable,
  colback=white, colframe=black!40,
  boxrule=0.5pt, arc=1mm
]
\subsection*{Execution and Retrieval}

\vspace{0.5em}
\noindent \textbf{Output format:}

\noindent Choose exactly ONE action and return exactly one JSON object (no markdown):

\noindent 1) RUN one coherent shell command:
\begin{verbatim}
{
  "action":"RUN",
  "reason":"...",
  "mode":"INSPECT|MUTATE|TEST|SERVICE|OTHER",
  "command":"...",
  "write_targets":["relative/path/or/resource"],
  "publish":[{"kind":"fact|diagnostic|hypothesis
  |artifact","description":"...",
  
  "evidence_steps":[1],
  
  "tags":["..."]}]
}
\end{verbatim}

\noindent For INSPECT/TEST use write\_targets=[] unless the command intentionally mutates persistent state. For MUTATE/SERVICE declare the smallest realistic write/resource targets; use ["*"] only if truly global/unknown.

\noindent 2) SEARCH reusable artifacts:
\begin{verbatim}
{"action":"SEARCH","reason":"...",

"query":"...","publish":[]}
\end{verbatim}

\end{tcolorbox}

\begin{tcolorbox}[
  enhanced, breakable,
  colback=white, colframe=black!40,
  boxrule=0.5pt, arc=1mm
]
\subsection*{Evidence-Grounded Decomposition}

\noindent 3) DECOMPOSE this node into one new AND-route:
\begin{verbatim}
{
  "action":"DECOMPOSE",
  "reason":"evidence-grounded reason
  this node should grow children",
  "evidence_steps":[12],
  "route_label":"short approach name",
  "children":[
    {
      "local_id":"c1",
      "goal":"one self-contained narrower obligation",
      "depends_on":[],
      "acceptance":[{"description":""}],
      "read_scope":["..."],
      "write_scope":["..."],
      "tags":["..."]
    }
  ],
  "publish":[]
}
\end{verbatim}

\noindent Child depends\_on may reference local\_id values from this same route or already-visible dependency node IDs. Do not reference the parent.

\end{tcolorbox}

\begin{tcolorbox}[
  enhanced, breakable,
  colback=white, colframe=black!40,
  boxrule=0.5pt, arc=1mm
]
\subsection*{Verification and Control Actions}

\noindent 4) VERIFY this node. If the current acceptance contract is missing/incomplete, propose up to 4 non-destructive checks:
\begin{verbatim}
{
  "action":"VERIFY",
  "reason":"why evidence suggests 
  
  the node may be complete",
  "checks":[{"description":"...","command":"..."}],
  "publish":[]
}
\end{verbatim}

\noindent Mechanical checks are preferred. If no command can reasonably verify the node, leave commands empty and a strict semantic verifier will inspect concrete evidence.

\noindent 5) ABANDON only the CURRENT METHOD when it is unproductive. This does NOT mean the obligation is impossible and does not fail the node:
\begin{verbatim}
{"action":"ABANDON","reason":"":[12],"publish":[]}
\end{verbatim}

\noindent 6) BLOCK only when concrete evidence shows this obligation cannot currently proceed under its contract/environment (e.g. unavailable required external resource). BLOCK is stronger than ABANDON and causes the containing route to fail so that a parent may try an alternative route:
\begin{verbatim}
{"action":"BLOCK","reason":"",
"evidence_steps":[12],
"publish":[]}
\end{verbatim}

\noindent 7) YIELD when another READY node should proceed or you have no useful next local action:
\begin{verbatim}
{"action":"YIELD","reason":"...","publish":[]}
\end{verbatim}

\noindent Never invent command outputs. Do not repeat an unchanged failed command.
Keep reason under 120 characters. Use publish=[] unless a claim will help later. Keep each command focused and JSON complete.

\end{tcolorbox}

\begin{tcolorbox}[
  enhanced, breakable,
  colback=white, colframe=black!40,
  boxrule=0.5pt, arc=1mm
]
\subsection*{Local Input Context}

\vspace{0.5em}
\noindent \textbf{Input Information (user message):}
\begin{verbatim}
LOCAL NODE CONTEXT:
{serialized_local_node_context}
\end{verbatim}
\noindent The serialized context contains the following runtime fields:
\begin{itemize}[leftmargin=*]
  \item \texttt{root\_task}: the original benchmark instruction.
  \item \texttt{node}: identifier, goal, depth, status, attempts, verification
  failures, acceptance conditions, scopes, tags, and runtime feedback.
  \item \texttt{parents}, \texttt{dependencies}, and
  \texttt{routes\_from\_this\_node}: relevant graph state and verified
  dependency artifacts.
  \item \texttt{local\_evidence} and \texttt{local\_artifacts}: recent
  execution records and outputs of the assigned node.
  \item \texttt{retrieved\_artifacts} and \texttt{last\_search\_query}:
  shared results retrieved by the runtime or an explicit search.
\end{itemize}
\end{tcolorbox}

\end{document}